\documentclass{svproc}
\usepackage{url}
\usepackage{graphicx}%

\usepackage{booktabs}%
\usepackage{amsmath,amssymb,amsfonts}%
\usepackage[utf8]{inputenc}
\usepackage[T1]{fontenc}
\usepackage[arabic,english]{babel}
\usepackage{array}
\begin{document}
\mainmatter              
\title{A Severity-Based Curriculum Learning Strategy
for Arabic Medical Text Generation}
\titlerunning{Severity-Based Curriculum Learning for Arabic Medical Text Generation}  
%

\author{
Ahmed Alansary \and
Molham Mohamed\and
Ali Hamdi
}

\institute{
October University for Modern Sciences \& Arts, Egypt\\
\email{ahmed.mohamed406@msa.edu.eg, molham.mohamed@msa.edu.eg, ahamdi@msa.edu.eg}
}

\maketitle              

\begin{abstract}
Arabic medical text generation is increasingly needed to help users interpret symptoms and access general health guidance in their native language. Nevertheless, many existing methods assume uniform importance across training samples, overlooking differences in clinical severity. This simplification can hinder the model’s ability to properly capture complex or high-risk cases. To overcome this issue, this work introduces a \textit{Severity-based Curriculum Learning Strategy for Arabic Medical Text Generation}, where the training process is structured to move gradually from less severe to more critical medical conditions. The approach divides the dataset into ordered stages based on severity and incrementally exposes the model to more challenging cases during fine-tuning, allowing it to first learn basic medical patterns before addressing more complex scenarios. The proposed method is evaluated on a subset of the Medical Arabic Question Answering (MAQA) dataset, which includes Arabic medical questions describing symptoms alongside corresponding responses. In addition, the dataset is annotated with three severity levels (Mild, Moderate, and Critical) using a rule-based method developed in this study. The results demonstrate that incorporating severity-aware curriculum learning leads to consistent performance improvements across all tested models, with gains of around +4\% to +7\% over baseline models and +3\% to +6\% compared with conventional fine-tuning approaches.

\keywords{Curriculum Learning, Arabic Medical Text Generation, Severity-based Training, Large Language Models, Natural Language Processing.}
\end{abstract}

\section{Introduction}
Medical text generation and understanding are increasingly central to modern healthcare, where clinical documents such as electronic medical records (EMRs), radiology reports, and physician notes contain essential information about diagnoses, treatments, and patient outcomes. These records support both clinical decision-making and large-scale medical analysis. However, working with medical text remains challenging due to its unstructured nature, specialized terminology, and wide variation in complexity. Clinical descriptions can range from routine observations to rare and highly complex cases, making automated processing difficult for natural language processing (NLP) systems \cite{paper17}. Moreover, generating reliable medical text requires models to capture not only frequent clinical patterns but also infrequent yet critical abnormalities, further increasing the complexity of the task \cite{paper13}.

Advances in deep learning and large language models have led to substantial progress in NLP, including medical text generation and understanding. Architectures such as recurrent neural networks, convolutional neural networks, and transformer-based models have been widely adopted in this domain. Through large-scale pretraining and contextual representation learning, these models can capture complex linguistic structures and domain-specific knowledge from extensive corpora \cite{paper17}. Despite these improvements, most training approaches still rely on uniform sampling, where all training instances are treated equally during optimization. This assumption does not align well with medical data, which are often highly imbalanced. Common conditions dominate clinical datasets, while rare or abnormal cases are underrepresented. Consequently, models tend to produce generic outputs that reflect frequent patterns, while struggling to accurately represent clinically important but infrequent cases \cite{paper13}. In addition, ignoring differences in sample difficulty may limit the model’s ability to learn more complex medical scenarios effectively \cite{paper17}.

Curriculum learning offers a structured alternative by introducing training samples in a meaningful sequence, typically moving from simpler to more complex examples. Inspired by human learning, this approach allows models to build foundational knowledge before tackling more difficult instances. Prior work has shown that such strategies can improve convergence behaviour and overall performance by guiding optimization more effectively in the early stages of training \cite{paper1}. In text generation, the order of training data has been shown to influence both learning efficiency and output quality, highlighting the potential of curriculum-based methods for neural generation tasks \cite{paper1}. In the context of medical NLP, recent studies have begun to explore curriculum learning as a way to better handle the inherent complexity and imbalance of clinical data, particularly for rare and challenging cases \cite{paper17}.

Building on these insights, this work proposes a severity-based curriculum learning strategy for Arabic medical text generation. Instead of treating all samples uniformly, the approach organizes training data according to levels of medical severity and introduces them progressively during fine-tuning. The model is first exposed to simpler cases, allowing it to learn basic medical patterns, before gradually incorporating more complex and severe conditions. This structured training process helps the model better capture the layered nature of medical knowledge and improves its ability to generate accurate and contextually appropriate outputs across a wide range of clinical scenarios.

\section{Related Work}
Natural language processing has become increasingly important in healthcare due to the growing availability of digital medical data such as electronic medical records, clinical notes, and online medical discussions \cite{paper9,paper19,paper20}. These resources enable automated systems to assist in tasks such as diagnosis support, medical question answering, and report generation. However, medical text processing remains challenging because clinical language is complex, domain-specific, and often unstructured \cite{paper11}. In addition, the biomedical domain contains specialized terminology and limited annotated datasets, which makes it difficult for general language models to perform reliably without domain adaptation \cite{paper12}. Recent studies have therefore explored adapting large language models to medical applications through fine-tuning, domain-specific corpora, and specialized training strategies.

Several works have focused on Arabic medical language processing due to the scarcity of resources and the linguistic complexity of the Arabic language. The rich morphology of Arabic, the diversity of dialects, and the limited availability of annotated datasets create additional challenges for developing reliable medical NLP systems \cite{paper10}. To address these issues, researchers have proposed various approaches such as dataset construction, ensemble learning, and specialized preprocessing techniques. For example, ensemble transformer models have been applied to Arabic biomedical question classification to improve prediction stability and accuracy \cite{paper5}. Other studies investigated text preprocessing techniques such as summarization, refinement, and named entity recognition to improve disease classification and symptom severity prediction from Arabic medical texts \cite{paper4,paper6}. Additionally, large language models have been fine-tuned on real-world medical conversations to generate medical advice and assist patients in healthcare systems \cite{paper7}. These approaches demonstrate the potential of transformer-based models in Arabic medical applications.

Another research direction focuses on intelligent healthcare systems that combine classification and generation tasks to assist medical decision-making. For example, transformer-based systems have been developed for medical triage and severity prediction using symptom descriptions provided by patients \cite{paper8}. Open-domain question answering systems have also been proposed to provide reliable health information in Arabic by integrating knowledge extraction and language generation models \cite{paper18}. In addition, synthetic data generation has been explored to address the scarcity of clinical data in privacy-sensitive domains \cite{paper14}.  These studies highlight the increasing role of generative models in healthcare applications.

Curriculum learning has emerged as an effective training strategy for improving neural models by organizing the learning process from easier samples to more difficult ones. Previous research demonstrated that curriculum learning can improve both convergence speed and model performance in data-to-text generation tasks \cite{paper1}. In the medical domain, curriculum-based frameworks have been applied to complex tasks such as medical report generation, where models progressively learn from simpler training instances before handling more challenging cases \cite{paper13}.  Curriculum learning has also been applied to other generation tasks such as emotional dialogue systems and machine translation to improve training stability and output quality \cite{paper15,paper16}.

Despite these advances, most current medical NLP systems still rely on standard training strategies that treat all samples equally during model optimization. Such approaches may not effectively capture the varying difficulty of medical cases or the imbalance between common and rare conditions. Motivated by these limitations, this work explores a curriculum-based training strategy that organizes medical data according to difficulty levels and progressively introduces more complex cases during model training.

\begin{table}
\centering
\caption{Example instances from the Arabic medical severity-aware dataset with English translations.}
\label{tab:dataset_examples}
\scriptsize
\begin{tabular}{p{0.45\textwidth} p{0.45\textwidth} p{0.10\textwidth}}
\hline
\textbf{Question} & \textbf{Answer} & \textbf{Severity}\\
\hline

\foreignlanguage{arabic}{ورم في ارقبه كيف اتعامل معه هل يستدعي جراحه} &
\foreignlanguage{arabic}{علي حسب مكانه ونوعه روح لطبيب اورام} &
\foreignlanguage{arabic}{حرج} \\

A neck lump, how should I deal with it? Does it require surgery? &
It depends on its location and type; consult an oncologist. &
Critical \\

\hline

\foreignlanguage{arabic}{اعاني من انتفاخ الخد نتيجه التورم اللثه بسبب تسوس الاسنان الماميه} &
\foreignlanguage{arabic}{يحتاج الامر لعلاج بمضاد حيوي وعمل علاج عصب} &
\foreignlanguage{arabic}{غير حرج} \\

I suffer from cheek swelling due to gum inflammation caused by front tooth decay. &
The condition requires antibiotic treatment and a root canal procedure. &
Mild \\

\hline

\foreignlanguage{arabic}{دايما احس بالم في الصدر من الجهه اليسار فوق الثدي} &
\foreignlanguage{arabic}{امرض القلب لاتعاطي الم مستمر ودائم احتمال يكون عندك شد عضلي اي تقلص في عضلات الصدر راجعي اي طبيب للفحص عليك} &
\foreignlanguage{arabic}{متوسط} \\

I always feel pain in the left side of my chest above the breast. &
Heart diseases do not usually cause constant pain; it may be a muscular strain. Consult a doctor for examination. &
Moderate \\

\hline
\end{tabular}
\end{table}

\section{Dataset: Severity Annotation}
This study utilizes the \textit{Medical Arabic Question Answering (MAQA)} dataset, which contains Arabic medical questions and their corresponding answers. The dataset was originally designed for open-domain medical question answering and includes textual pairs consisting of a medical \textit{question} describing symptoms and a corresponding \textit{answer} that provides medical advice or guidance. In this work, a subset of 32K question–response pairs from the dataset was used for training and evaluation.

Since the original dataset does not provide annotations related to the medical urgency of the questions, an additional severity annotation was introduced to support the proposed \textit{severity-based curriculum learning strategy}. Specifically, each question was assigned one of three severity levels: \textit{Mild}, \textit{Moderate}, or \textit{Critical} as illustrated in table \ref{tab:dataset_examples}. These labels represent the estimated medical urgency of the symptoms described in the question and enable the organization of the training data into progressively more difficult learning stages.

The annotation process was implemented using a rule-based approach based on a curated list of Arabic medical symptom keywords. First, the questions were preprocessed using basic Arabic text normalization, including the removal of diacritics and punctuation, as well as the normalization of common letter variations. After preprocessing, each question was analyzed for the presence of symptom-related keywords associated with different levels of medical severity. Questions containing terms related to emergency symptoms, such as severe chest pain, loss of consciousness, severe bleeding, or difficulty breathing, were labeled as \textit{Critical}. Questions describing intermediate conditions such as fever, vomiting, or persistent pain were labeled as \textit{Moderate}. Questions containing mild symptoms, including headache, common cold, or minor discomfort, were labeled as \textit{Mild}.

If multiple keywords appeared in the same question, the highest severity level was assigned according to the following priority order: \textit{Critical} $>$ \textit{Moderate} $>$ \textit{Mild}. Questions that did not match any predefined symptom keywords were assigned the \textit{Mild} label by default.

The resulting severity annotations enable the dataset to be organized into staged training subsets corresponding to increasing levels of medical complexity. This structure directly supports the proposed curriculum learning framework, in which the model is first trained on mild cases and is progressively exposed to moderate and critical cases during fine-tuning.

\begin{figure*}[t]
    \centering
    \includegraphics[width=0.7\linewidth]{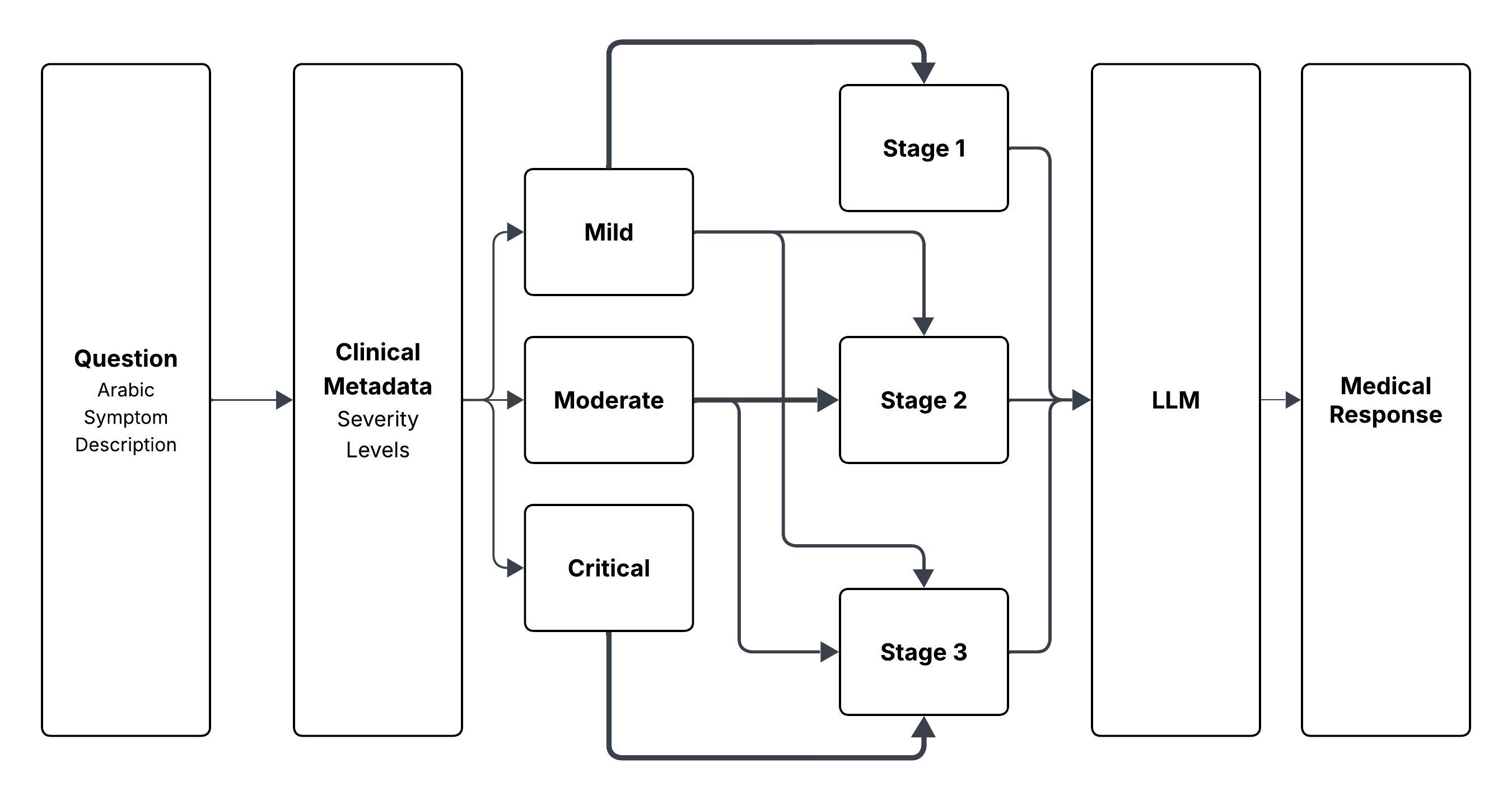}
    \caption{Overview of the proposed severity-based curriculum learning framework}
    \label{fig:method_overview}
\end{figure*}

\section{Methodology}
The proposed a framework aims to improve the ability of language models to generate informative medical responses by organizing the training process according to the medical severity of the cases. Instead of training the model using a uniform sampling strategy, the learning process progressively exposes the model to increasingly complex medical cases. In this way, the model first learns basic medical response patterns from mild cases before adapting to more complex and high-risk medical situations. The overall framework of the proposed severity-based curriculum learning strategy is illustrated in Fig.~\ref{fig:method_overview}.

\subsection{Problem Formulation}

Let $\mathcal{D}=\{(q_i,a_i,s_i)\}_{i=1}^{N}$ denote the medical question–response dataset, where $q_i$ represents a medical question describing symptoms, $a_i$ denotes the corresponding medical response, and $s_i \in \{Mild, Moderate, Critical\}$ indicates the severity level assigned during the annotation process described in Section~III. The goal of the model is to generate an appropriate medical response $\hat{a}_i$ given an input question $q_i$.

The generation model is trained in a conditional language modeling setting, where the probability of generating a response sequence $a=(y_1,\dots,y_T)$ is defined as

\[
P(a|q)=\prod_{t=1}^{T} P(y_t \mid y_{<t}, q; \theta),
\]

where $\theta$ denotes the model parameters and $y_{<t}$ represents previously generated tokens. The training objective is to minimize the negative log-likelihood of the reference responses across the dataset:

\[
\mathcal{L}_{CE}=-\frac{1}{N}\sum_{i=1}^{N}\sum_{t=1}^{T_i} \log P(y_t^{(i)} \mid y_{<t}^{(i)}, q_i; \theta).
\]

\subsection{Severity-based Curriculum Learning}
To incorporate the notion of medical severity into training, the dataset is divided into three subsets based on annotated severity levels:

\[
\mathcal{D}_{mild}, \quad \mathcal{D}_{moderate}, \quad \mathcal{D}_{critical}.
\]

Each subset reflects an increasing level of medical complexity. A curriculum learning approach is then adopted, where the model is fine-tuned in stages, gradually exposing it to more severe cases.

At the initial stage, training is restricted to mild cases only:

\[
\mathcal{D}_1 = \mathcal{D}_{mild}.
\]

This step enables the model to capture fundamental medical language patterns and general response structures associated with simpler and more common conditions.

In the second stage, moderate cases are introduced:

\[
\mathcal{D}_2 = \mathcal{D}_{mild} \cup \mathcal{D}_{moderate}.
\]

Here, the model learns to handle more detailed and accurate medical descriptions, while retaining the knowledge acquired from the earlier stage.

In the final stage, the full dataset is utilized by including critical cases:

\[
\mathcal{D}_3 = \mathcal{D}_{mild} \cup \mathcal{D}_{moderate} \cup \mathcal{D}_{critical}.
\]

This gradual progression supports learning from simpler to more challenging scenarios, improving the model’s ability to manage complex and severe medical situations.

\subsection{Model Fine-tuning}

The curriculum strategy is applied to a set of pretrained Arabic language models using parameter-efficient fine-tuning. In particular, Low-Rank Adaptation (LoRA) is used to update the transformer weights while leaving the majority of parameters unchanged. This setup reduces computational requirements while preserving the knowledge gained during pretraining.

For training, each input is constructed as a single sequence consisting of a question followed by its corresponding answer. The model is optimized to generate the answer conditioned on the question using the standard cross-entropy objective defined earlier. The curriculum framework determines the sequence in which training samples are presented.

\subsection{Training Procedure}

Training is carried out in three consecutive stages aligned with the curriculum levels. At each stage, model parameters are updated using the corresponding dataset partition. To ensure stable learning as task difficulty increases, the learning rate is gradually decreased across stages. Let $\theta_k$ represent the model parameters after stage $k$. The optimization at each stage is given by

\[
\theta_k = \arg\min_{\theta} \mathbb{E}_{(q,a) \sim \mathcal{D}_k} \left[\mathcal{L}_{CE}(q,a;\theta)\right], \quad k \in \{1,2,3\}.
\]

This stepwise optimization allows the model to progressively adapt from simpler to more complex medical cases.

Overall, the methodology combines severity-aware data organization with curriculum-based fine-tuning, facilitating structured knowledge acquisition and enhancing the generation of coherent and informative Arabic medical responses.

\section{Results and Discussion}
Table~\ref{tab:extended_results} reports the performance of baseline models, standard fine-tuning, and the proposed curriculum learning strategy across a range of Arabic language models. Several clear patterns emerge regarding how curriculum-based training affects medical text generation.

\begin{table*}[!htbp]
\centering
\footnotesize
\renewcommand{\arraystretch}{0.5}
\caption{Performance comparison across Baseline, Standard Fine-tuning, and Curriculum Learning.}
\label{tab:extended_results}
\begin{tabular}{l c c c}
\toprule
\textbf{Model} 
& {\textbf{Baseline}} 
& {\textbf{Standard Fine-Tuning}} 
& {\textbf{Curriculum Learning}}  \\
\midrule
AraGPT2-Base        & 54.04\% & 59.71\% & 63.21\%  \\
AraGPT2-Medium      & 59.16\% & 60.14\% & 64.38\%  \\
Bloomz-560M         & 63.59\% & 63.65\% & 65.57\% \\
Al-Atlas-0.5B       & 61.03\% & 61.52\% & 65.65\% \\
GPT2-Small-Arabic   & 56.40\% & 58.88\% & 61.02\%  \\
LFM2-350M           & 63.34\% & 63.72\% & 67.02\% \\
LFM2-700M           & 63.46\% & 63.88\% & \textbf{68.00\%} \\
Dialect-ar-gpt-2021 & 60.41\% & 60.73\% & 65.63\% \\
Qwen2.5-0.5B   & 57.83\% & 57.96\% & 64.46\% \\
Qwen3-0.6B     & 60.43\% & 61.40\% & 64.65\%  \\
\bottomrule
\end{tabular}
\end{table*}

Standard fine-tuning leads to only modest gains over the baseline in most cases. Improvements are typically limited to around 0.06\%–2\%, suggesting that conventional fine-tuning does not substantially alter how the models learn from the data. Since all training samples are treated uniformly, the models are not explicitly guided to handle differences in complexity or clinical severity, which may limit learning efficiency.

By contrast, the curriculum learning strategy yields consistent and more pronounced improvements across all models. Relative to the baseline, performance increases fall roughly in the +4\% to +7\% range. Even when compared against standard fine-tuning, additional gains of about +3\% to +6\% are observed. This indicates that ordering the training data by difficulty contributes meaningfully to more effective learning.

The extent of improvement differs across model sizes. Smaller and medium-sized models benefit the most, with gains often reaching +5\% to +7\% over the baseline. This pattern suggests that curriculum learning provides useful structure during training, allowing models with limited capacity to first capture simpler patterns before progressing to more complex medical cases.

\begin{figure*}[h]
\centering
\includegraphics[width=0.8\linewidth]{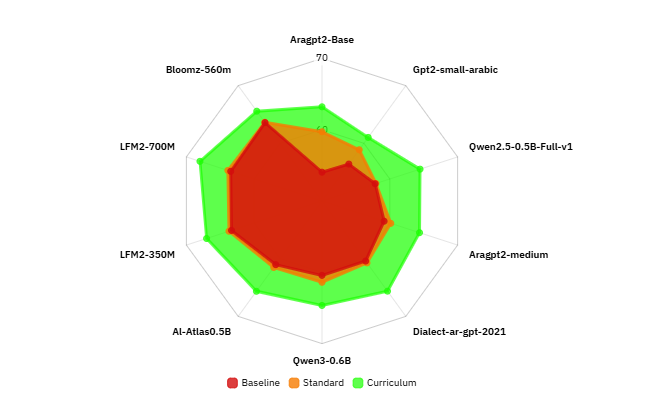}
\caption{Visual comparison of model performance across baseline, standard fine-tuning, and curriculum learning approaches.}
\label{fig:model_improvements}
\end{figure*}

Larger models also show consistent improvements, although their baseline performance is already relatively high. In these cases, gains are generally within +3\% to +5\%. While smaller in magnitude, these improvements are sufficient to produce the highest overall scores. Notably, the LFM2-700M model achieves the best result at 68.00\%, indicating that curriculum learning can further refine strong models by improving optimization efficiency.

Another observation is the uniformity of the improvements across different architectures. All models listed in Table~\ref{tab:extended_results} show performance gains under curriculum learning, regardless of size or design. This suggests that the advantage stems primarily from how the training data are organized, rather than from architecture-specific factors.

Figure~\ref{fig:model_improvements} visualizes these trends. A consistent upward shift is visible when curriculum learning is applied, with average gains of approximately +4\% to +6\% across models. This supports the idea that gradually introducing more complex samples helps models better capture medical language patterns.

Overall, structuring the training process by sample difficulty leads to clear and consistent performance improvements. Beyond increasing final accuracy, curriculum learning also promotes a more stable and efficient training process when dealing with the variability inherent in Arabic medical text.

\section{Conclusion}
This work introduced a method that targets a key shortcoming of conventional fine-tuning approaches, where all training instances are treated uniformly despite notable differences in medical complexity and urgency. By structuring the training data according to severity levels and gradually incorporating more complex cases throughout fine-tuning, the framework allows the model to first acquire essential medical response patterns before addressing more demanding and critical scenarios.

To enable this strategy, a severity annotation scheme was developed for the MAQA dataset using a rule-based approach grounded in medically relevant symptom keywords. These annotations made it possible to divide the dataset into staged subsets corresponding to mild, moderate, and critical cases, thereby facilitating the application of the proposed curriculum learning process. Consequently, training was carried out as a multi-stage optimization procedure, where the model was incrementally exposed to increasingly complex medical examples.

The experimental findings indicate that severity-aware curriculum learning enhances both the quality and consistency of generated Arabic medical responses when compared with standard fine-tuning methods. The approach supports the model in capturing the hierarchical nature of medical knowledge and improves its capacity to handle varied symptom descriptions.

In summary, the results emphasize the value of incorporating medical severity into the training of generative models for healthcare applications. The proposed framework offers a straightforward and effective means of improving Arabic medical text generation and can potentially be adapted to other medical NLP tasks in which case difficulty and clinical urgency differ across data samples.

\bibliographystyle{splncs04}
\bibliography{templates/ref}

\end{document}